# MACHINE LEARNING MODEL SURVEY WITH THE DATASET FOR SOLAR FLARE PREDICTION

Erik Larsen

PeopleTec, Inc., 4901 Corporate Dr., Huntsville, AL 35805
erik.larsen@peopletec.com

## ABSTRACT

*High energy solar flares and coronal mass ejections have the potential to destroy Earth's ground and satellite infrastructures, causing trillions of dollars in damage and mass human suffering. This would lead to food shortages and crippled emergency response capabilities. A solution to this impending problem is proposed herein using satellites in solar orbit with built-in machine learning capability that continuously monitor the Sun. They will use machine learning to calculate the probability of massive solar explosions from the remote sensing data, then signal defence mechanisms that can mitigate the threat. This paper reports the results from a survey of machine learning models using open-source solar flare prediction data. The rise of edge computing supports machine learning hardware placed on the same satellites as the sensor arrays, saving critical transmit time across the vast distances of space. A system of systems approach will allow enough warning for safety measures to be enacted, thus mitigating the risk of disaster.*

## 1. INTRODUCTION

A powerful Class-X solar flare accompanied by a Coronal Mass Ejection (CME) – a cascade of highly energetic particles accelerated from the sun's corona by magnetic field collapse – could cause catastrophic damage to Earth's ground electronic and orbital satellite infrastructures. These giant eruptions have the capability to wreak havoc on GPS and other satellites, airplane communications, power grids, copper wiring in transformers [1], and even hand-held modern devices like smart phones. Time will be of the essence when a Class-X event occurs. While complete negation of the threat may be impossible with today's technology, prior knowledge of the event will provide time to enact protection protocols; this temporary approach is similar to preparing for a hurricane. Due to the increased reliance of our society on satellites and electronics for essential communication, transportation, and commerce, a solution to protect the global infrastructure must be implemented to prevent widespread calamity and human strife.

A system of systems involving remote sensing and machine learning could mitigate this disaster. The top-level requirement of this approach is spaced-based solar remote sensing platforms with on-board machine learning (ML) capability. These satellites will constantly monitor Sol, processing and feeding collected data to algorithms trained in calculating the probability of highly energetic flares and CMEs. Once the probability crosses a certain threshold, protective measures – e.g., putting satellites into a sleep mode – can be implemented. A more advanced defensive shield would involve swarms of satellites with nanomaterials tuned to absorb and reflect the incoming radiation being deployed at an interdicting position once the signal is received. Ultimate success will require pre-positioning the satellites at all stages well in advance to give adequate warning and preparation time.

This work presents analysis of the Data Set for Solar Flare Prediction (DSFP) [2] and a survey of machine learning algorithms trained on it to calculate the probability of a massive solar flare using the PyCaret module [3]. The best performing algorithms should be optimized with a rigorous hyper-parameter grid search, maximizing prediction accuracy while minimizing false negatives (FN) due to the disastrous consequences of misclassifying a positive solar flare event. Understanding and synthesis of the data will be invaluable as an inevitable solar flare catastrophe nears [4]. Advance warning given by this method will allow the appropriate agencies to take protective action with enough time to minimize damage to ground infrastructure and orbital satellites.

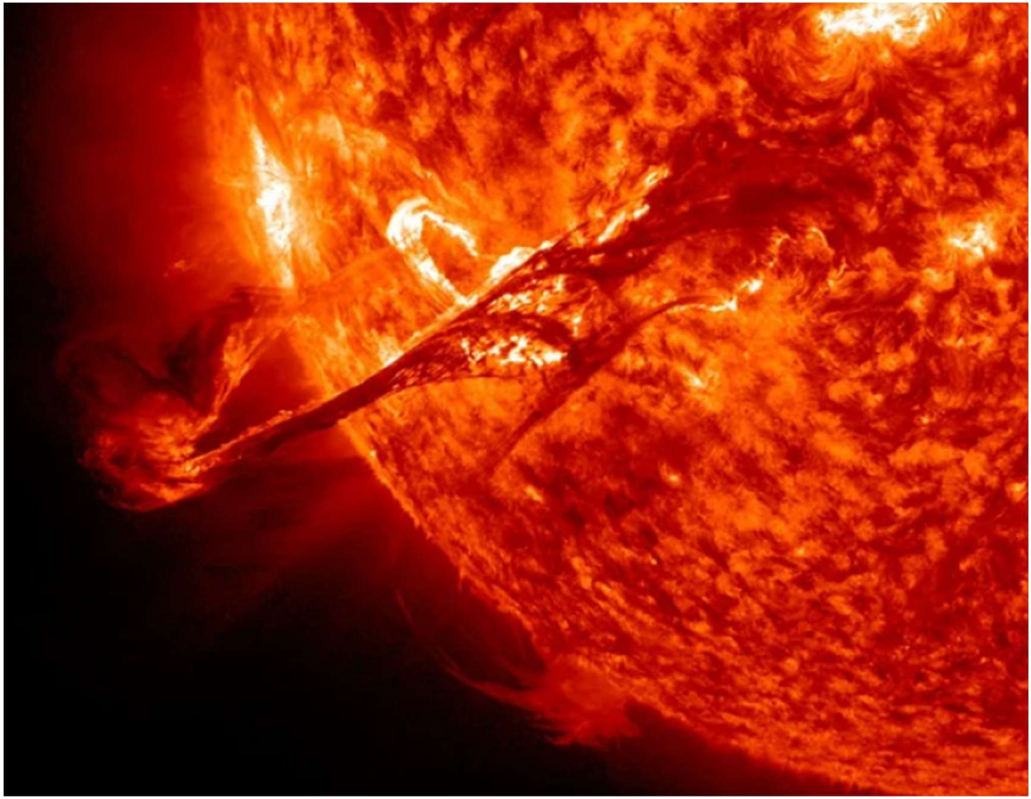

Figure 1. Solar flare captured by NASA's Solar Dynamics Observatory in August 2012

## 2. LITERATURE REVIEW

### 2.1. Current Understanding of Solar Flares

In order to predict solar flares with ML algorithms it is necessary to understand their cause so that relevant and proper data collection occurs. Sol is an active star whose magnetic motion drives constantly changing radiative, electromagnetic, and particulate outputs [5]. Sunspots are caused by dense magnetic field lines protruding through the photosphere causing a cool area to form relative to their surroundings at 5800K. Magnetohydrodynamics from the rotating core drive these phenomena, which are exacerbated by the difference in rotation speed between the Sun's layers, equator, and poles. Flares are thought to be triggered by the reconnection of these magnetic field lines [6].

This activity "varies over timescales ranging from the very short to the very long – stellar and planetary evolutionary timescales" [5]. Sunspots follow what is called the solar cycle with a period of 11 years during which these magnetically induced surface dark and light spots occur displaying maximum and minimum frequency. Flares are often associated with CMEs that carry a vast amount of charged, magnetized plasma out into the solar system at high velocities [7].

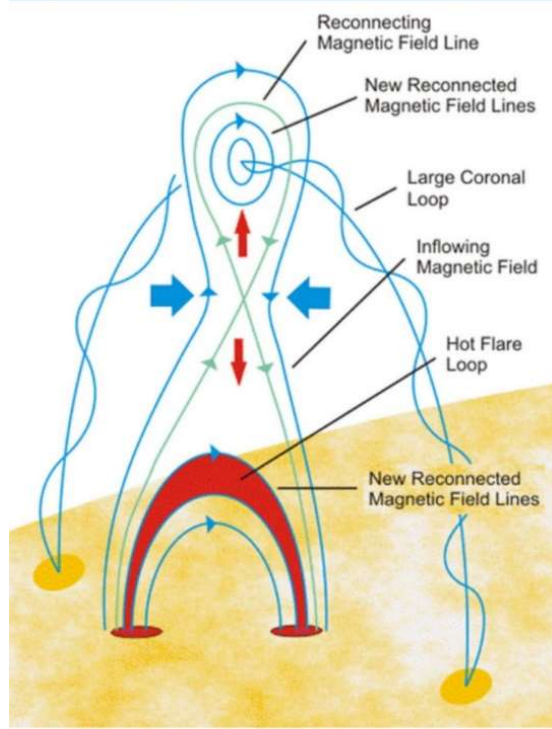

Figure 2. Diagram of the standard model of solar flares and magnetic field reconnection. Image credit: Gordon Holman and NASA.

## 2.2. Historical Context

The most infamous flare on record is the Carrington Event of 1859 which caused telegraph communications around the world to fail and a global scale aurora that could be seen in tropical islands like Cuba and Jamaica [8]. Named after an amateur astronomer – Richard Carrington – it was discovered while observing sunspots with his personal telescope. This flare was so powerful that it caused telegraph machines to spark. One telegraph manager reported that "platinum contacts were in danger of melting" [8]. The induced auroras glowed so brightly that it seemingly turned night into day: birds began chirping and people began their workday chores before realizing it was the middle of the night. While some thought the world was ending, the true cause was the storm of energetic particles strewn from the sun by a Class-X solar flare releasing as much energy as 10 billion atomic bombs. Ice core samples have helped determine that the Carrington event was twice as large as all the other solar storms in the last 500 years. If an event of this scale were to happen today the damage is estimated to be in the trillions of dollars [1].

A lesser-known example of an extremely powerful solar flare is the so-called New York Railroad Storm in 1921 [4]. Love et al. [9] described this storm as being comparable in magnitude to the 1859 Carrington event, as measured by the storm time disturbance index, or Dst. Due to the lack of more sophisticated instruments and record keeping abilities in the mid-19th century, it is difficult to accurately gage the Carrington storm's exact magnitude. Measured in nanoteslas (nT), the maximum Dst an intense storm can register is several hundred nTs. The Carrington superstorm is believed to register between -850 – -1050 nT according to the sparse data of the era available to analysts. This uncertainty makes it difficult to compare with modern storms, whose Dst index has been kept in much greater detail since 1957 [9]. Looking into the nature of these magnetic superstorms, the researchers perused observatory records and found the 1921 storm to be approximately -976 nT (Figure 3); this puts the 1921 event on a scale close to the 1859 event, both happening in a timespan of less than 100 years.

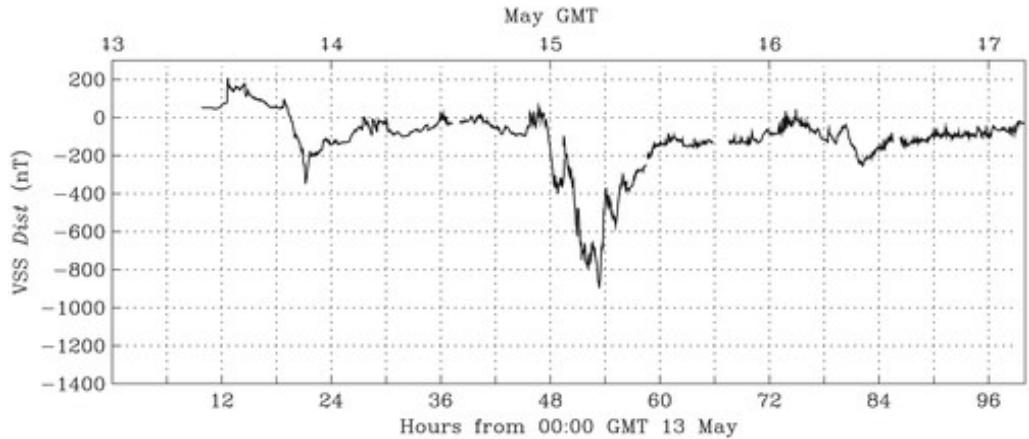

Figure 3. Vassouras (VSS) horizontal magnetic field component readings from May 1921

Prior to the magnetic field disturbances, sunspots were observed in a complex, near-equatorial group which rotated west from the Sun's east limb; spectroheliograms also showed considerable activity in and around the group [9]. Such a causal relationship lends credence to the idea that catastrophic storms can be predicted with machine learning. Considering that this event occurred over 100 years ago, it is reasonable to project that another tremendous solar event is likely to occur in the near future.

In addition to the aforementioned damage, [10] revealed that a space weather event in August of 1972 detonated dozens of U.S. Navy sea mines after reaching Earth in only 14.6 hours. The researchers determined that the ejecta was directed at Earth and produced by a sunspot causing brilliant flares which cleared a path for the fast-moving particles. Much like the Carrington event, effects from this solar blast were noticed around the world [11]. There was damage to solar panels on orbiting satellites as well as false nuclear bomb detonation signals from Air Force sensors being switched on. Astronauts en route to the moon (or at least outside of Earth's protective envelope) would be in imminent danger from the radiation. Events such as these have led communities dependent on space weather to research how these storms could impact today's infrastructure. While it is well known that light takes approximately 8 minutes to reach Earth from the sun, advanced warning of several hours could prove sufficient to enact basic precautions.

## 3. ENABLING TECHNOLOGICAL ADVANCES

Earth's defense is a topic that has been studied for decades and is not just relegated to the purview of solar disasters. The idea of a solution to catastrophic events from asteroid impacts has been researched extensively by [12], and even includes the establishment of a Planetary Defense Coordination Office (PDCO). This organization has discovered over 98% of the nearly 16,000 near Earth objects (NEO) on catalog since its inception. While plans to mitigate an extinction level event are considered, they also emphasize that detection with remote sensing devices is the first critical step. It is conceivable that a similar body could be convened for protection from solar based threats which incorporates elements of the detection and prediction technology reviewed herein.

### 3.1. Solar Remote Sensing Technology

Research into the building blocks of this solution have already begun. The rise of fossil fuels has sparked a growing interest in green energy, and hence in forecasting solar radiation for solar energy gathering systems dependent on accurate data. For example, the Autoregressive Integrated Moving Average (ARIMA) model is used with the combination of regression-base ML and sensor data to make such predictions [13]. Remote sensing of the Sun has been taking place at NASA for decades by missions including the Solar & Heliospheric Observatory (SOHO), Solar Dynamics Observatory (SDO), and Advanced Composition Explorer (ACE) [14, 15, 16] discussed in the following sections. Table 1 summarizes the equipment being used on these platforms.

Table 1. Solar Remote Sensing Missions & Instruments

| Mission | Instrument | Target Measurements |
| --- | --- | --- |
| SOHO | Global Oscillations at Low Frequency (GOLF) | Global Sun velocity oscillations |
| | Variability of solar Irradiance and Gravity Oscillations (VIRGO) | Low resolution imaging; active cavity radiometers |
| | Michelson Doppler Imager / Solar Oscillations Investigations (MDI/SOI) | Velocity oscillations, harmonic degree to 4500 |
| | Solar Ultraviolet Measurements of Emitted Radiation (SUMER) | Normal incidence spectrometer: 50 – 160 nm |
| | Coronal Diagnostic Spectrometer (CDS) | Normal and grazing incidence spectrometer: 15 – 80 nm |
| | Extreme-Ultraviolet Imaging Telescope (EIT) | Full disk imager; chromospheric and coronal structures |
| | Ultraviolet Coronagraph Spectrometer (UVCS) | Coronal diagnostic between 1.3 and 10 solar radii |
| | Large Angle Spectroscopic Coronagraph (LASCO) | Triple white light coronagraph between 1.1 and 30 solar radii |
| | Solar Wind Anisotropies (SWAN) | Telescope for Lyman-α absorption |
| | Charge, Element, and Isotope Analysis System (CELIAS) | Solar wind energy between 0.1-1000 keV/e |
| | Comprehensive Super Thermal and Energetic Particle Analyzer (COSTEP) | Solare wind ion energy between 0.04 – 53 MeV/n |
| | Energetic and Relativistic Nuclei and Electron Experiment (ERNE) | Solar wind and isotope composition |
| SDO | Helioseismic and Magnetic Imager (HMI) | High resolution full disk; vector magnetograms |
| | Atmospheric Imaging Assembly (AIA) | Solar atmosphere; 10 different wavelengths |
| | Extreme Ultraviolet Variability Experiment (EVE) | Extreme UV spectral irradiance |
| ACE | Electron, Proton, and α-particle Monitor (EPAM) | Energetic ions and electrons |
| | Magnetic Field Monitor (MAG) | Magnetic field vectors |
| | Solar Isotope Spectrometer (SIS) | High energy particle flux |
| | Solar Wind Electron, Proton, and Alpha Monitor (SWEPAM) | Solar wind ions |

### 3.1.1. SOHO

The Solar and Heliospheric Observatory mission was launched in December of 1995 as a joint endeavor between NASA and the European Space Agency (ESA). Its assignment is to study the inner and outer structure of Sol, including the corona and even solar winds. Among its myriad discoveries, SOHO gathered the first images of the Sun's convection zone and consequently the sub-surface structure of sunspots. It is equipped with multiple remote sensing platforms. During its observations SOHO has discovered over 3,000 comets and phenomena like solar tornadoes and coronal waves. There are 12 instruments giving a wide range of sensing capabilities, from composition and individual wavelengths of light, to "measuring energetic particles passing the spacecraft" [14].

### 3.1.2. SDO

The Solar Dynamics Observatory was designed to facilitate understanding of Sol's influence on Earth by "studying the solar atmosphere on small scales of space and time and in many wavelengths simultaneously" [15]. It was launched on February 11, 2010 and has furthered our understanding of how space weather is produced by solar activity. SDO is equipped with three remote sensing experiments that simultaneously observe the Sun: Atmospheric Imaging Assembly (AIA), EUV Variability Experiment (EVE), and the Helioseismic and Magnetic Imager (HMI). They can take rapid-fire snapshots in super high-definition of everything from solar flares, magnetic activity, and even the solar dynamo located deep below the surface; this produces an incredible 1.4 terabytes of data every day from its geosynchronous orbit [15].

### 3.1.3. ACE

The Advanced Composition Explorer – a.k.a. Explorer 71 – collects particles as it orbits the sun. These can be of solar, interplanetary, interstellar, and inter-galactic origins. It provides space weather reports and gives warnings of geomagnetic storms harmful to satellites and astronauts in space. Located at the Sun-Earth L1 Lagrange point, ACE was launched on August 25, 1997 and continues to operate, possibly until 2024. It boasts nine instruments varying from mass spectrometers to ionic charge analyzers and magnetometers [16].

## 3.2. Machine Learning

The advent of graphical processing units (GPU) has engendered the rise of ML. Technology miniaturization has brought high-speed computing into edge devices that are connected in a new web called the Internet-of-Things [17]. These advances have crept into our lives in the form of Siri or Alexa, recommendation systems found in popular applications such as Netflix, LinkedIn, or Facebook, and even smart home appliances like refrigerators and thermostats. ML algorithms are capable of providing suggestions for new friends or movies based on a user's past behavior [18]; they can even detect when you are low on milk and seamlessly order more for you from a grocery store. Such advances make possible the placement of ML capable hardware directly on future solar observation satellites, further reducing event prediction time by eliminating signal transit time over astronomical distances.

Machine learning is already being used to forecast solar weather and understand solar life-cycles. [19] ask the pertinent question: "are there reliable ways of detecting these events?" To answer it, the research team constructed a convolutional neural network (CNN) to detect solar events using only magnetograms from non-events. Data was collected from the Joint Science Operations Center (JSOC) at Stanford University, including readings from the SDO and SOHO missions discussed above [20]. Figure 4 displays a sample magnetogram from the Stanford Solar Center (SSC). While they were only able to achieve 50% test accuracy (i.e., results comparable to guessing), they conclude a system with more robust data and less computational limitations will make a reliable detection system possible.

[21] advanced this idea by using Support Vector Machines (SVM) and multilayer perceptrons (MLP) along with decision tree algorithms to predict flares greater than Class-M and Class-C. Using vector magnetic field data taken from the HMI onboard the SDO satellite and the Space-Weather Heliospheric and Magnetic Imager Active Region Patches (SHARPs) data they make predictions of three classes for active regions (AR) that lead to solar eruptions. These three identified subsets are flares, flares associated with CMEs and solar energetic particles (SEP), and

CME only. They use data standardization with median and standard deviation values. The investigators use a one-vs-all approach with the SVM as it is designed for binary classification problems. "The SVM then classifies the data by placing a separating hyperplane with the maximum distance between the classes of the data" [21].

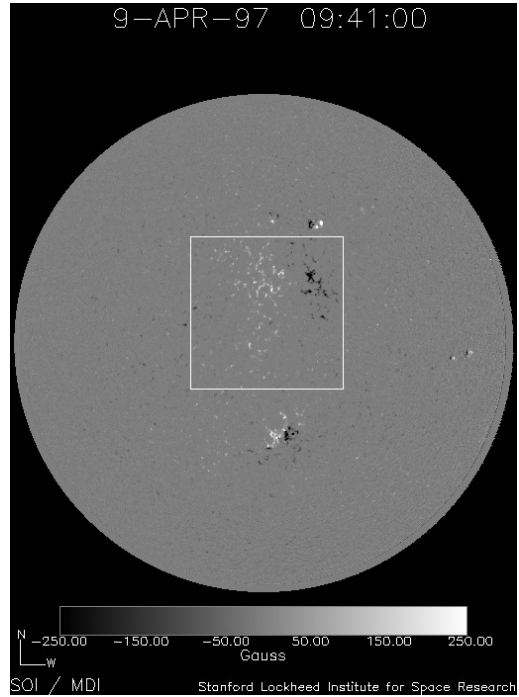

Figure 4. Example magnetogram for the SSC taken during solar minimum

The MLPs produce a posterior probability that the ingested data belongs to one of the classes. This is akin to a multinomial logistic regression where the hidden layers are activated by the prolific logistic activation function and the final output layer uses a softmax activation. Various numbers of hidden layer neurons – from 18 to 108 in steps of 18 – are tested with a hyper-parameter grid search for precise tuning. They employ stratified k-fold cross-validation techniques to alleviate errors from bias, variance, and class population differences. Optimization of the weight parameters is accomplished with the popular L-BFGS algorithm [21]. To assess performance the True Skill Statistics are used. This metric is simply the probability of false detection (POFD) subtracted from the probability of detection (POD). These, in turn, are taken from the true positive (TP), false positive (FP), true negative (TN), and false negative (FN) values of the resulting confusion matrix.

Their principal finding is that "SVMs can successfully predict if a flare will be accompanied with CMEs and SEPs 96 hours prior to their occurrence" [21] with a TSS value of $0.92 \pm 0.09$. They also show that the SVM model can predict if a CME will not be associated with a flare or SEP for longer prior time intervals with a TSS above $0.90 \pm 0.10$. MLPs have similar performance, predicting that a flare will not be associated with CMEs or SEPs 96 hours prior to its occurrence with a TSS value of $0.91 \pm 0.07$. These findings show that SVMs typically perform slightly better than MLPs in forecast windows of 96 and 36 hours with consistent TSS values above 0.90; in short, "SVMs are slightly better than MLPs" [21] for this classification task.

Two years later, [22] further advanced this area of research by including temporal information in the model architecture. The construction of a recurrent neural network (RNN) model called Long-Short Term Memory (LSTM) captures temporal information by adding inputs from previous calculations to those further down the feed forward path. Their research goal is to predict if an active region (AR) will produce a highly energetic flare within 24 hours. The time evolution of photospheric magnetic field patterns could serve as triggering indicators of CMEs as well as flares. Wang's research group uses the aforementioned SHARP data to train their ML model along with the Geostationary Operational Environmental Satellite (GOES) X-ray flare catalogs to identify and correlate flare events. The TSS value is calculated as previously discussed to assess performance of the LSTM on this data. Standardization occurs using mean and standard deviation values, optimization with the powerful "Adam" optimizer, and a learning rate of 0.001 [22].

Regularization to combat bias and variance is implemented as well, and the model is trained for only six epochs after which there is no significant score improvement. The LSTM outperforms previous attempts by Inceoglu et al. using SVMs and MLPs. The introduction of non-linearity reduces the number of FPs while boosting TP accuracy. Their research concludes that the LSTM model is a valid method for forecasting solar flares [22]. Prediction accuracy and TSS scores can be improved by incorporating solar cycle activity knowledge and more data. Together all of these reports show conclusively that solar flares can be predicted with ML algorithms using data collected by solar remote sensing instruments.

## 4.0. Data Analysis and ML Model Comparison Findings

The most important insights gleaned from this paper will be an optimal machine learning algorithm and the feature space derived from the data with which to train said algorithm. Exploratory Data Analysis (EDA) illuminates important sensor reading correlations and eliminates certain others showing redundancy. The following sections detail the dataset, performance metrics, and model comparisons used. Calculations are made on a Dell 7550 laptop with a 16 GB NVIDIA Quadro RTX5000 GPU, and in online Kaggle.com notebooks listed in the References section [23, 24]. This work is coded in the Python programming language with scikit-learn's preprocessing packages for standardization and scaling. Knowledge gained from the results will lay the foundation for future endeavors that lead to a proof-of-concept, eventual satellite deployment, and ultimately help provide enough warning to minimize harmful impact from highly energetic flares, CMEs, and SEPs.

### 4.1. Dataset for Solar Flare Prediction (DSFP)

Hollanda et al. made the DSFP available on June 9$^{th}$, 2021 through publication by Elsevier. The authors created this new set by compounding records of magnetic attributes and solar flare data. They used Python's SunPy library to gain access to JSOC, GOES, and SWPC archives. Table 2 contains several sample features and their descriptions; the full feature space consists of 31 different variables whose explanations can be found in [2]. It contains 8,874 samples from the period between May 2010 and December 2019. There are 8493 non-flare entries and 381 flare samples designated as 0 and 1, respectively. The SHARP data is recorded every 12 minutes – a smaller temporal resolution than SDO – for each AR 24 hours prior to a flare. Observations were filtered to lie within $\pm 70°$ of the Sun's central meridian for noise reduction. They use a z-score based standardization method, then execute 5-fold-based training and test set splitting for robustness [2].

Table 2. A Sample of DSFP Features.

| Attribute Name | Description |
| --- | --- |
| FLARE_NUMBER | Represents occurrence of a solar flare or not with values of 1 corresponding to M- or X-class events |
| QUALITY | A flag from SHARP representing quality of the data as related to noise |
| TOTBSQ | Total magnitude of the Lorentz force |

| TOTPOT | Total photospheric magnetic free energy density |
|--------|------------------------------------------------|
| MEANGBT | Average gradient of the total magnetic field |
| MEANJZD | Average vertical current density |

### 4.1.1 Exploratory Data Analysis (EDA)

A first look at the data shows it uses over 2.1 MB of memory, and all entries are of type float64 or int64 except T_REC, which is a datetime object – i.e., the date and time of recording. A check for missing values reveals 15 null entries under the MEANSHR category, which reports the mean shear angle. Eleven of these values belong to negative events, leaving four missing values corresponding to positive flare occurrences. There is a multitude of ways to handle missing data: removing examples with missing entries, dropping the MEANSHR category altogether, and various forms of imputation to fill in the blanks. Since this missing data only accounts for 1.05% of positive examples and 0.13% of negative examples these entries are simply dropped from the bulk before further analysis.

Graphical visualizations of the DSFP show overlapping positive and negative class distributions in many features, a few of which are seen in Figure 5. There is some separation in target distribution for the R_VALUE feature, but most positive instances – those labeled as 1 – simply exhibit much tighter grouping within the spread of negative examples. Without further feature engineering this can limit performance of ML models by increasing the false positive and, more importantly, false negative rates.

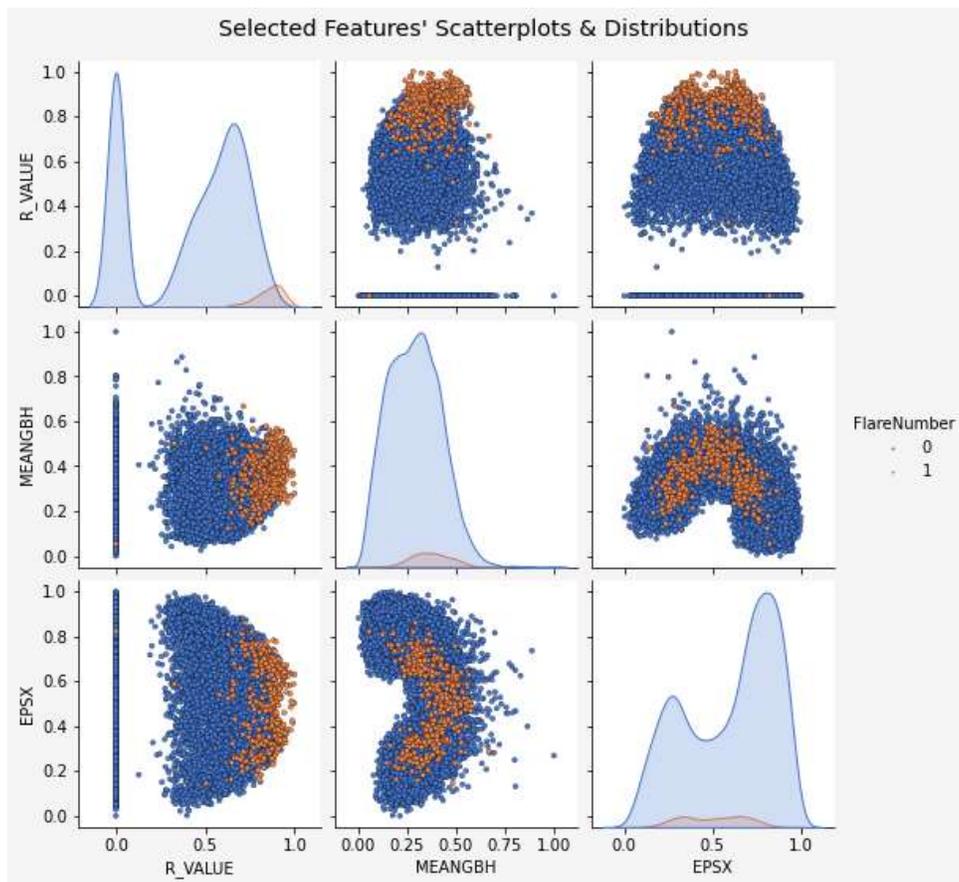

Figure 5. Overlapping and grouping of feature values

After standardization the data is more evenly distributed. Figure 6 displays a heatmap of the Pearson R correlation for each of the non-standardized features. The picture reveals that there are several highly correlated features. Highly correlated variables contain redundant information. For this reason, one of a pair of such variables can sometimes be removed from the dataset before model ingestion. This reduces the overall complexity and computation time, while often increasing prediction accuracy. However, removing these features is not accompanied by the desired performance increase. Therefore, all DSFP features are included. Figure 7 shows the plotted relationships between a few of these.

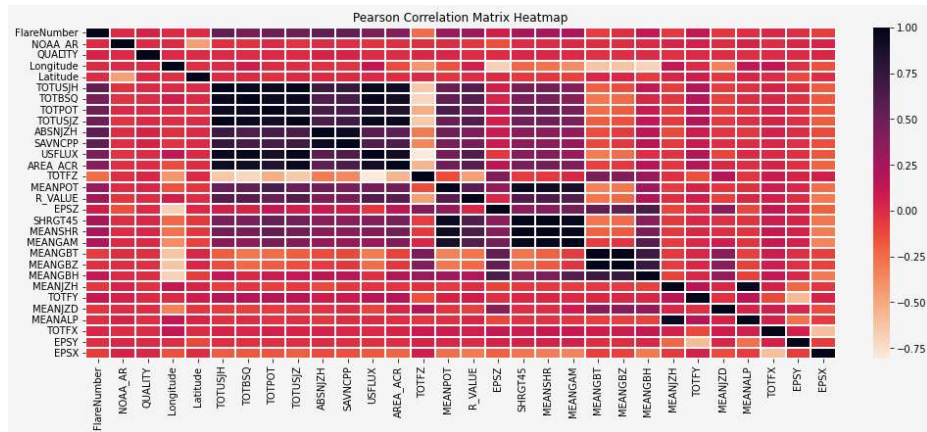

Figure 6. DSFP Feature Correlations with Unscaled Data

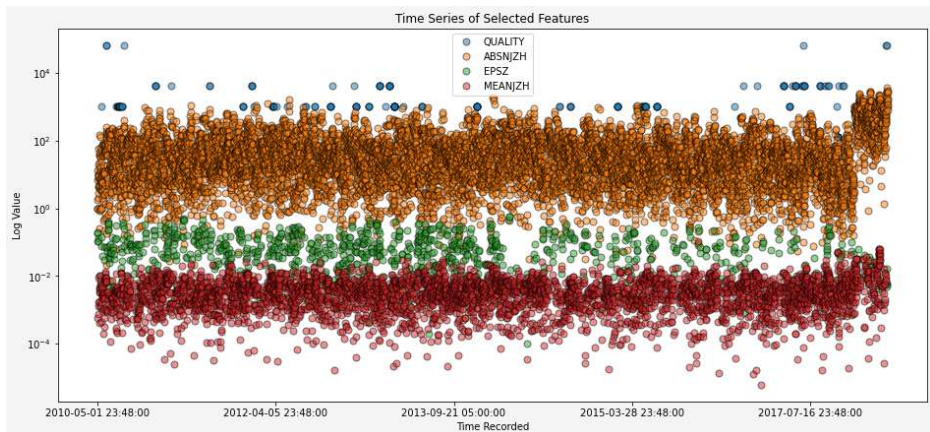

Figure 7. Log plot shows higher values after 2017

### 4.1.2. Pycaret ML Model Comparison

PyCaret is a high-level, open-source, Python-based ML library. It quickly compares multiple algorithms using the given data then ranks them according to metrics that include accuracy, F1 score, and the Matthews Correlation Coefficient (MCC) among others. PyCaret has many functions built-in to pre-process data such as normalization using the z-score and population imbalance correction using the Synthetic Minority Oversampling Technique (SMOTE) [3].

Accuracy is not always the best metric by which to judge the success of ML models. Precision and recall must be taken into account when there are dire consequences for misclassification. A common metric used is the F1 score which considers false positives and false negatives as well.

Table 3 displays the results of PyCaret's model comparison using the DSFP data. Models are ranked according to accuracy. In the top ranks there are many decision tree-based methods, with CatBoost Classifier (catboost) scoring the highest in accuracy, and only Linear Discriminant Analysis achieves a slightly higher F1 score. Equation 3 shows that increasing false negatives decreases the F1 score. This vital metric incorporates both precision (P) and recall (R) according to the following equations:

$$P = \frac{TP}{TP + FP} \quad (1)$$

$$R = \frac{TP}{TP + FN} \quad (2)$$

$$F1 = 2 \cdot \frac{(P \cdot R)}{P + R} = \frac{TP}{TP + \frac{1}{2}(FP + FN)} \quad (3)$$

Table 3. PyCaret Model Comparison Results.

| Model | Accuracy | AUC | Recall | F1 | MCC | TT (Sec) |
|---|---|---|---|---|---|---|
| Cat Boost Classifier | 0.9703 | 0.9512 | 0.4626 | 0.5624 | 0.5808 | 8.5500 |
| Logistic Regression | 0.9690 | 0.9486 | 0.4255 | 0.5278 | 0.5514 | 0.7500 |
| Ridge Classifier | 0.9686 | 0.0000 | 0.3414 | 0.4626 | 0.5105 | 0.1790 |
| Extreme Gradient Boosting | 0.9676 | 0.9342 | 0.4405 | 0.5315 | 0.5472 | 0.5460 |
| Light Gradient Boosting Machine | 0.9673 | 0.9363 | 0.4442 | 0.5352 | 0.5536 | 1.2280 |
| Linear Discriminant Analysis | 0.9671 | 0.9375 | 0.5282 | 0.5723 | 0.5705 | 0.0670 |
| Random Forest Classifier | 0.9658 | 0.9345 | 0.3526 | 0.4502 | 0.4833 | 0.7820 |
| Extra Trees Classifier | 0.9652 | 0.9305 | 0.3303 | 0.4276 | 0.4598 | 1.3120 |
| Gradient Boosting Classifier | 0.9632 | 0.9355 | 0.4295 | 0.5020 | 0.5096 | 4.5850 |
| Ada Boost Classifier | 0.9602 | 0.9181 | 0.3966 | 0.4410 | 0.4560 | 0.9840 |
| K Neighbors Classifier | 0.9597 | 0.8042 | 0.2935 | 0.3691 | 0.3930 | 0.4060 |
| Decision Tree Classifier | 0.9431 | 0.6785 | 0.3884 | 0.3794 | 0.3621 | 0.1500 |
| Quadratic Discriminant Analysis | 0.9111 | 0.9379 | 0.7992 | 0.4766 | 0.4901 | 0.0330 |
| Naïve Bayes | 0.9024 | 0.9143 | 0.7365 | 0.4278 | 0.4351 | 0.0180 |
| SVM – Linear Kernel | 0.6213 | 0.0000 | 0.9669 | 0.1865 | 0.2402 | 7.0760 |

The F1 score for each model is much lower than accuracy but is higher than that found in previously mentioned works of Inceoglu and Wang. While high accuracy is promising and false positives are essentially benign, the misclassification of an impending catastrophic flare event will cause global calamity. Population balancing with the SMOTE method and highly collinear feature removal have a detrimental effect on these test scores. This is likely due to the larger number of non-correlated features, where useful information is lost when these features are removed. Minority oversampling also decouples the positive events from the time series aspect, negating any positive effects that could be gained from sequential data. The F1 score is critical because it considers the number of false negatives, and area under the curve (AUC) is a measure of false positive rate versus the true positive rate. The number of FNs is greater than the number of FP as seen in Figure 8, but both contribute to lowering the F1 score and should be minimized.

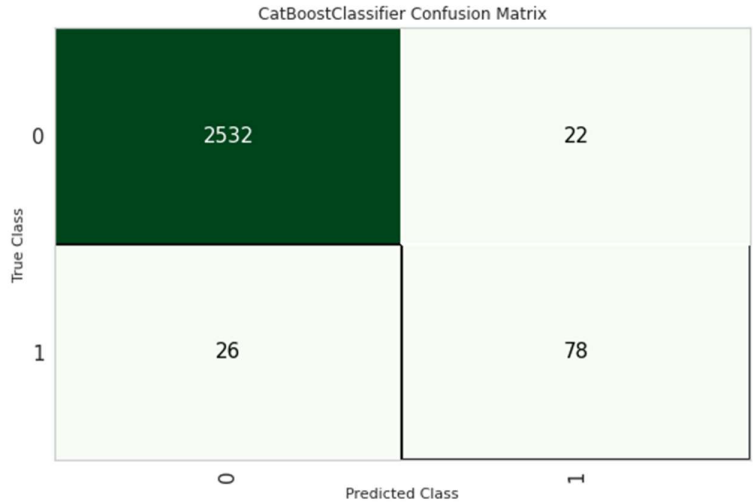

Figure 8. Correct and incorrect classifications on test data

The Receiver Operator Characteristic (ROC) curve is shown in Figure 9. The AUC is very high, meaning that there is a very high TP rate when the FP rate is small. This shows that Cat Boost Classifier is much more successful at predicting when a solar event will occur than simply guessing (shown by the dotted black line where $AUC = 0.50$). Figure 10 shows that Longitude and R_VALUE are the most influential features.

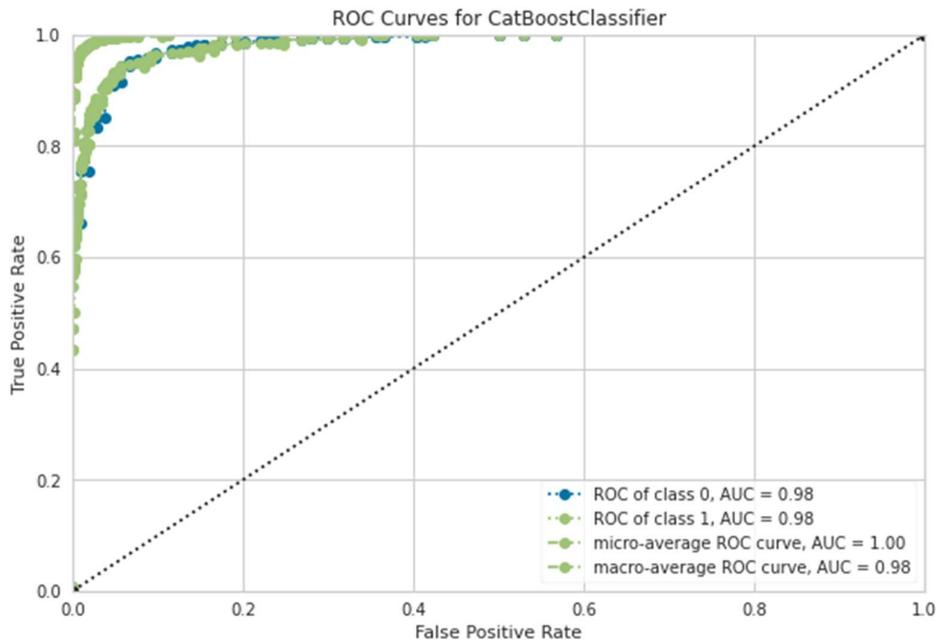

Figure 9. ROC curve and AUC scores

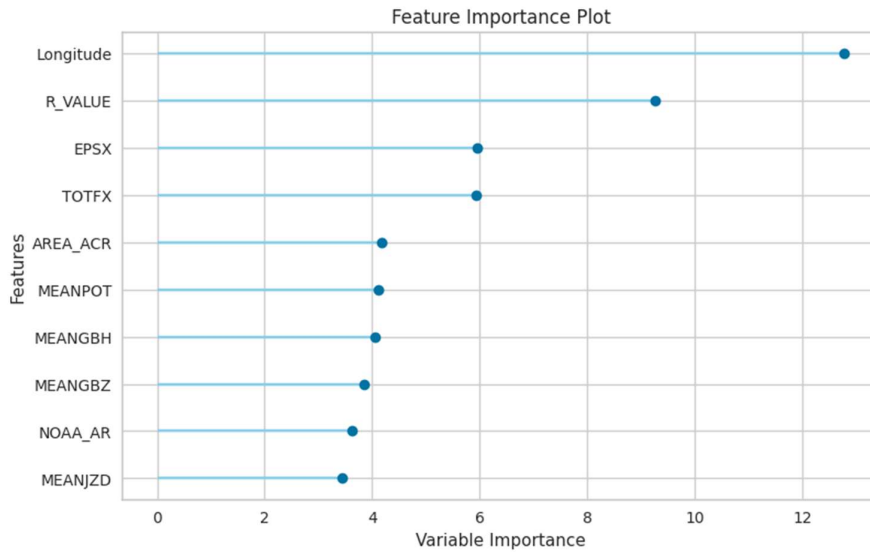

Figure 10. Most influential features

## 5.0. CONCLUSIONS

This research builds upon the work of previous groups to advance the survivability of Earth's satellite, electrical and communications infrastructures from a highly energetic solar flare event. To accomplish this, fifteen known machine learning algorithms are surveyed identifying the best model for near-term deployment. Recently released data called the Dataset for Solar Flare Prediction is used to train these models in the task of predicting a catastrophic solar flare with high true positive accuracy and minimized false negatives. Results support the hypothesis that machine learning algorithms can help predict solar flares and hence should be included in plans for defending infrastructure against them.

The best performing model in a PyCaret comparison is the Cat Boost Classifier, followed by Logistic Regression. Results are consistent with previous works with a slight improvement of the F1 score. Data collected from solar remote sensing satellites is critical for this method. Hence, the most advanced instruments should be used to detect the physical features that are deemed most influential to the ML model's ability to accurately make predictions. While accuracy is high for most models, the F1 score shows a vulnerability: a high number of false negatives leading to decreased recall and F1 scores.

Future research must involve choosing hardware platforms for deployment. This includes a satellite mission with advanced remote sensing capabilities focused on mid-range solar longitudes, and the type of processors necessary for durable extended operation in the solar system at less than one astronomical unit. Analysis must be conducted to decide how many satellites should be deployed to cover the Sun's surface. A network of satellites providing continuous coverage that is redundant, flexible, and evolvable should be deployed as detailed by Vladimirova et al. (2008). Novel combinations such as RADAR and LIDAR on the same vessel should also be considered. Falconi et al. (2021) have shown that a photonic integrated circuit (PIC) built with silicon on insulator (SOI) technology is such a viable platform. Such a combination exploits the different strengths of both remote sensing technologies.

Further algorithm testing should include ensemble methods similar to model stacking, with an emphasis on improving the recall score. Ensemble algorithms take the output of the individual models, giving each a weighted "vote." The final output is decided from this score and is often an improvement over the individual model predictions. In this way it may be possible to capitalize on the strengths of each model while bolstering the weak areas to improve recall. Recurrent neural networks like LSTM should be included in the decision process to introduce non-linearity that may further reduce misclassifications. Considering the possible outcomes of poor preparation, the cyclic nature of extreme solar activity, and time passed since the last major event, detailed planning for this mission should begin as soon as possible.

## ACKNOWLEDGEMENTS

The authors would like to thank Dr. Jonathan W. Campbell of Embry-Riddle Aeronautical University and the PeopleTec, Inc. Technical Fellows program for encouraging and assisting this research.

## AUTHORS

Erik Larsen, M.S. is a senior data scientist with research experience in quantum physics and deep learning. He completed both M.S. and B.S. in Physics at the University of North Texas, and a B.S. in Professional Aeronautics from Embry-Riddle Aeronautical University while serving as an aviator in the U.S. Army.

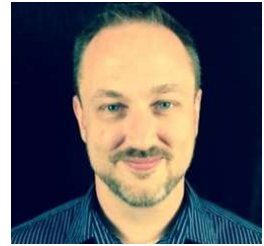